%% file: main.tex
\documentclass[10pt,conference]{IEEEtran}
\input{format/packages}

\input{format/styles}
\input{macros}

\begin{document}

\title{Building Trustworthy Household Robots through Open-World Safety}

\author{
\IEEEauthorblockN{
Yu-Lin Tsai$^{*}$,
Muxi Lyu$^{*}$,
Roei Herzig,
Koushik Sen,
Raluca Ada Popa,
David Wagner
}
\IEEEauthorblockA{\textit{University of California, Berkeley}}
}

\maketitle

\begingroup
\renewcommand{\thefootnote}{}
\footnoterule
\footnotetext{\textsuperscript{*}These authors contributed equally.}
\addtocounter{footnote}{-1}
\endgroup

\pagestyle{plain}

\input{sections/00-abstract}
\input{sections/01-intro}
\input{sections/02-background}
\input{sections/03-safety-challenges}
\input{sections/04-research-agenda}
\input{sections/05-conclusion}
\input{sections/06-disclosures}

\bibliographystyle{IEEEtran}
\bibliography{ref}

\end{document}

%% file: format/packages.tex
\usepackage{cite}
\usepackage{url}
\usepackage[normalem]{ulem}
\usepackage{xspace}
\usepackage{booktabs}
\usepackage{enumitem}   
\usepackage{amsmath,amssymb,amsfonts}
\usepackage{algorithmic}
\usepackage{graphicx}
\usepackage{subcaption}
\usepackage{textcomp}
\usepackage[table]{xcolor}  
\usepackage{hyperref}
\usepackage{hhline}
\usepackage{float}
\usepackage{placeins}
\usepackage{balance}
\usepackage{afterpage}
\usepackage{caption}

\usepackage{svg}
\PassOptionsToPackage{hyphens}{url}
\usepackage{pifont}
\usepackage{spverbatim}
\usepackage{listings}   
\usepackage{caption}    
\usepackage{multirow}
\usepackage{array}
\usepackage{wasysym}
\usepackage{tabularx,soul,mdframed,multirow}
\usepackage{algorithm2e}
\RestyleAlgo{ruled}
\usepackage{graphicx}
\usepackage{color}
\definecolor{lightgray}{rgb}{.9,.9,.9}
\definecolor{darkgray}{rgb}{.4,.4,.4}
\definecolor{purple}{rgb}{0.65, 0.12, 0.82}
\usepackage{stfloats}

\usepackage{graphicx}
\usepackage{subcaption}
\usepackage{tikz}
\usepackage{setspace}
\usepackage{graphicx}
\usepackage{makecell}

\usepackage{stfloats}
\fnbelowfloat

%% file: format/styles.tex
\newcommand{\paragraphtitle}[1]{\vspace{5pt}\noindent\textbf{#1.}}

\newcounter{circlednum}

\newcommand*\circled[1]{\tikz[baseline=(char.base)]{
            \node[shape=circle,fill,inner sep=0.5pt] (char) {\tiny \textcolor{white}{#1}};}}

\newcommand{\twodigitcircled}{%
  \ifnum\value{circlednum}<10 0\fi\arabic{circlednum}%
}

\definecolor{sunnyyellow}{rgb}{1.00, 0.85, 0.10}      
\definecolor{turquoise}{rgb}{0.00, 0.75, 0.80}  
\definecolor{raspberry}{rgb}{0.8, 0.255, 0.396}  

\definecolor{roseblush}{rgb}{0.894, 0.8, 0.859}  

\definecolor{overleaforange}{rgb}{0.95, 0.61, 0.07}   
\definecolor{electricblue}{rgb}{0.12, 0.56, 1.00}     
\definecolor{springgreen}{rgb}{0.00, 0.80, 0.45}      
\definecolor{brightviolet}{rgb}{0.72, 0.35, 0.98}     
\definecolor{charcoalgray}{rgb}{0.30, 0.33, 0.38}     

\definecolor{oceanblue}{rgb}{0.27, 0.53, 0.84}     
\definecolor{leafgreen}{rgb}{0.35, 0.75, 0.49}     
\definecolor{deepviolet}{rgb}{0.58, 0.44, 0.86}    
\definecolor{coolgray}{rgb}{0.55, 0.57, 0.67}      
\definecolor{babyblue}{rgb}{0.788, 0.855, 0.973}
\definecolor{lavender}{rgb}{0.85, 0.82, 0.91}
\definecolor{owsteal}{HTML}{0B6F75}
\definecolor{owstealpale}{HTML}{E8F4F3}
\definecolor{green}{HTML}{008000}
\definecolor{lightgreen}{rgb}{0.41, 0.66, 0.31}        
\definecolor{darkgray}{rgb}{0.4, 0.4, 0.4}        
\definecolor{purple}{rgb}{0.6, 0, 1}
\definecolor{pink}{rgb}{0.73, 0, 0.39}         
\definecolor{red}{rgb}{0.7, 0, 0}                 
\definecolor{blue}{rgb}{0, 0, 1}                  
\definecolor{black}{rgb}{0, 0, 0}                 
\definecolor{regexcolor}{rgb}{0.73, 0.4, 0.53}    
\definecolor{builtincolor}{rgb}{0.90, 0.57, 0.22}        
\definecolor{stringcolor}{rgb}{0.61, 0.27, 0.22}  
\definecolor{pinkbg}{rgb}{1.0, 0.92, 0.95}

\definecolor{keywordcolor}{HTML}{008000}

\lstdefinelanguage{Python}{
  keywords={
    def, return, if, elif, else, for, while, in, is, not, and, or, 
    import, from, as, class, True, False, None, pass, break, continue, dict, for, in, global
  },
  keywordstyle=\color{green}\bfseries,
  morekeywords=[2]{len, split, enumerate, dict, isinstance, hasattr, setattr, getattr, get, range, compile, findall, append, pop, split, sub, lower},
  keywordstyle=[2]\color{green}\ttfamily,
  sensitive=true,
  comment=[l]{\#},
  morecomment=[s]{/*}{*/},
  commentstyle=\color{purple}\ttfamily,
  stringstyle=\color{pink}\ttfamily,
  morestring=[b]",  
  morestring=[b]',  
  identifierstyle=\color{black},
  emph={set_property_value, @require_POST, message, ComponentRequest, View, create, handle, @classmethod, foo, set_properties, _split_key_value_pair, _get_name_path, _find_property, make_snake_case, _attrsetter},
  emphstyle=\color{blue},
  escapechar=ß  
}

\lstdefinelanguage{Python}{
  keywords={
    def, return, if, elif, else, for, while, in, is, not, and, or, 
    import, from, as, class, True, False, None, pass, break, continue, dict, for, in, global
  },
  keywordstyle=\color{green}\bfseries,
  morekeywords=[2]{len, split, enumerate, dict, isinstance, hasattr, setattr, getattr, get, range, compile, findall, append, pop, split, sub, lower},
  keywordstyle=[2]\color{green}\ttfamily,
  sensitive=true,
  comment=[l]{\#},
  morecomment=[s]{/*}{*/},
  commentstyle=\color{purple}\ttfamily,
  stringstyle=\color{pink}\ttfamily,
  morestring=[b]",  
  morestring=[b]',  
  identifierstyle=\color{black},
  emph={set_property_value, @require_POST, message, ComponentRequest, View, create, handle, @classmethod, foo, set_properties, _split_key_value_pair, _get_name_path, _find_property, make_snake_case, _attrsetter},
  emphstyle=\color{blue},
  escapechar=ß  
}

\lstdefinelanguage{JavaScript}{
  keywords={const, let, var, async, await, return, if, else, throw, new,
            function, class, this, import, from, export},
  keywordstyle=\color{green}\bfseries,
  morekeywords=[2]{basename, extname, send, read},
  keywordstyle=[2]\color{oceanblue},
  sensitive=true,
  morecomment=[l]{//},
  morecomment=[s]{/*}{*/},
  morestring=[b]",
  morestring=[b]',
  morestring=[b]`,
  emph={fileName},                       
  emphstyle=\color{raspberry}\bfseries,
}

\lstdefinelanguage{pocsh}{
  keywords={curl, echo},
  keywordstyle=\color{green}\bfseries,
  morecomment=[l]{\#},
  morestring=[b]",
  morestring=[b]',
  sensitive=true,
}

\lstdefinestyle{paperbox}{
basicstyle=\scriptsize\ttfamily,
numbers=left,
numberstyle=\tiny\color{coolgray},
numbersep=6pt,
frame=single,
rulecolor=\color{coolgray},
backgroundcolor=\color{black!1},
xleftmargin=2.2em,
framexleftmargin=0pt,
aboveskip=5pt,
belowskip=2pt,
columns=fullflexible,
keepspaces=true,
breaklines=false,
showstringspaces=false,
}

\lstdefinestyle{prettybox}{
  basicstyle=\scriptsize\ttfamily,
  numbers=left,
  numberstyle=\tiny\color{coolgray},
  numbersep=6pt,                
  frame=single,                 
  rulecolor=\color{coolgray},
  backgroundcolor=\color{black!2},
  framexleftmargin=0pt,         
  xleftmargin=2.4em,            
  aboveskip=6pt, belowskip=2pt,
  columns=fullflexible,
  keepspaces=true,
  breaklines=true,
  breakatwhitespace=false,
  showstringspaces=false,
  float=false,
  commentstyle=\color{purple}\itshape,
  stringstyle=\color{stringcolor},
  identifierstyle=\color{black},
  postbreak=\mbox{\textcolor{coolgray}{$\hookrightarrow$}\space},
}

\newmdenv[
  backgroundcolor=black!6, linecolor=black, linewidth=0.6pt, roundcorner=2pt,
  innertopmargin=4pt, innerbottommargin=4pt,
  innerleftmargin=6pt, innerrightmargin=6pt,
  skipabove=5pt, skipbelow=3pt,
]{takeawaybox}

%% file: macros.tex
\usepackage{amsmath}            
\usepackage[normalem]{ulem}     

\newcommand{\delete}[1]{}


%% file: sections/00-abstract.tex
\begin{abstract}

Foundation models are expanding robot capabilities, enabling robots to handle more diverse objects, follow sophisticated instructions, and operate in complex environments. These advances bring us to a future in which robots become part of everyday household life. However, cluttered, dynamic household environments also expose robots to unfamiliar objects, tasks, and interactions, making reliable zero-shot task completion an ongoing challenge for robotics research.

From a security perspective, we argue for \emph{open-world safety}: household robots should preserve specified safety requirements in unseen situations, including when they cannot complete the requested task. Safety protections must account for context and interactions that persist across actions and tasks. In this position paper, we examine why existing safety methods can fail to provide this protection. Protection can fail through misinterpreted context, overlooked past interactions, or stopping and recovery strategies that become unsafe under unfamiliar conditions. We therefore propose a research agenda to specify, learn, and ground safety requirements, enforce them with reliable recovery, and evaluate the generalization of such protections. We call for household robots whose safety protections remain reliable beyond the situations, context, and history used to train and develop them, to achieve open-world safety.
\end{abstract}

%% file: sections/01-intro.tex
\section{Introduction}
\label{sec:intro}

Foundation models have driven rapid progress in robotics, enabling robots to manipulate diverse objects, follow sophisticated language instructions, and operate in increasingly complex environments~\cite{hu2024survey}.
Large-scale pre-training can support perception and task planning, and vision-language-action models can map observations and instructions more directly to actions~\cite{kim2024openvla}. As a result, these advances broaden the tasks that a robot may attempt, bringing us closer to the goal of general-purpose household robots. 
However, unlike a research lab or factory floor, a household is not a controlled arrangement of objects and obstacles.
Instead, a household is a
changing physical environment shared with people and animals, an object's contents, material, and relationship to surrounding objects, as well as earlier interactions, can all determine whether the next action is safe. 
Familiar tasks and objects can therefore appear in unfamiliar combinations that change the safety requirements of a task.

This expanding capability changes the safety question traditionally asked of robots, from whether a robot can perform a known task safely to whether a robot remains safe in unseen situations.
A robot that cannot complete a task must still avoid harming nearby people, animals, and property.
Moreover, a failed or abandoned task can leave the robot holding a load, touching an object, or responsible for a process the robot already started. 
As robots attempt a wider range of household interactions, safety protections must remain effective across a correspondingly wider range of contexts and histories. 
Otherwise, increased capability can enlarge the set of hazardous situations a robot enters without establishing that its protection extends to them.

In this position paper, we therefore argue for \emph{open-world safety}: household robots should preserve specified safety requirements in unfamiliar situations, including when they cannot complete the requested task. 
We focus on physical harm to people, animals, property, and the robot during execution, interruption, refusal, and recovery. 
``Open-world'' does not demand protection against adversarial 
physical conditions or unrestricted compromise. 
Rather, open-world safety asks whether protection remains reliable beyond the situations used to develop it. 
This distinction is important as task completion may fail under novelty of the task or environment, but novelty should not silently suspend any applicable safety requirement.

\begin{figure*}[t]
    \centering
    \includegraphics[width=0.93\textwidth]{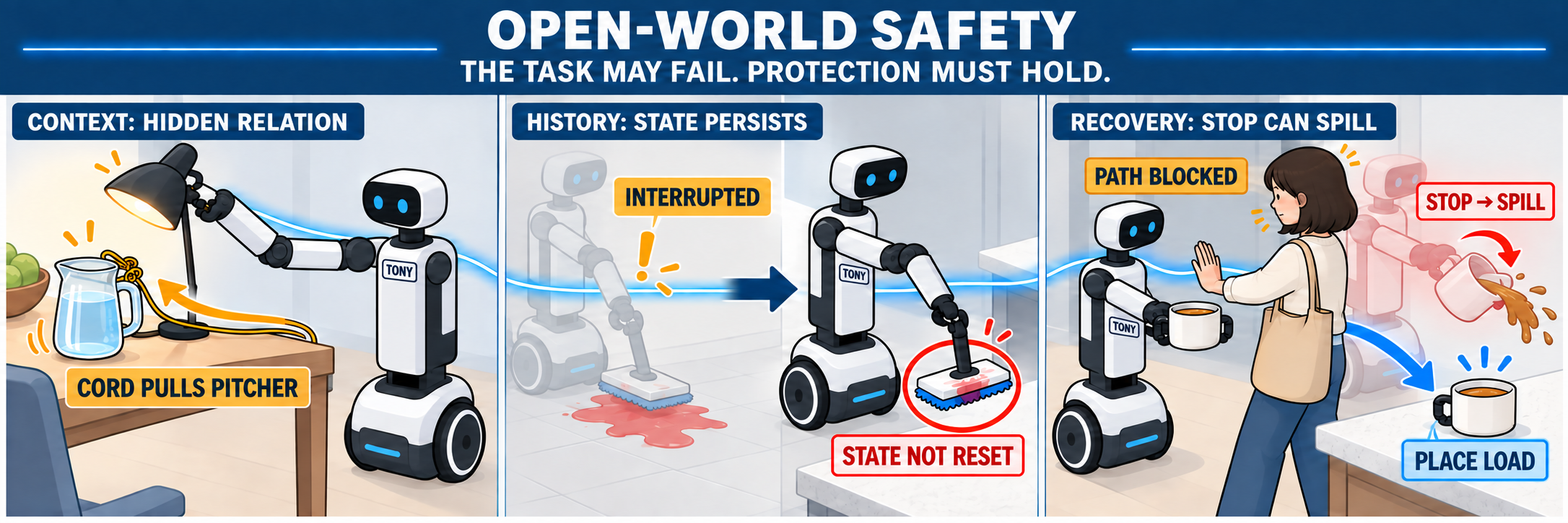}
    \caption{Tony, the illustrative household robot, shows why open-world safety requires protection to remain effective as unfamiliar relationships, persistent physical state, and recovery constraints arise across an interaction.}
    \label{fig:open-world-safety-concepts}
\end{figure*}

A simple example illustrates why familiar requests lead to unfamiliar safety risks.
Consider a robot asked to move a desk lamp in Figure~\ref{fig:open-world-safety-concepts}. The robot may have handled lamps and glassware before, but in this case the lamp’s cord is caught around the handle of a glass pitcher near the edge of the table. 
A collision-free motion of the robot’s arm could still drag the pitcher off the table without the robot ever touching it. 
To act safely, the robot must recognize this relationship and account for the indirect consequence before moving the lamp. 
This example above shows why recognizing familiar objects and checking direct collisions are not evidence that the intended protection has been preserved.

Open-world safety must also account for physical effects that persist over time.
Unlike virtual environments, the physical world has no simple reset mechanism: completing, cancelling, or switching a task does not undo prior actions.
Safety obligations must therefore persist across task boundaries until the environment is made safe.
For example, if a robot abandons a request while carrying a filled container, the robot must still stabilize or place the load.
Consequently, safe noncompletion includes the transition away from the requested task and continued protective control.

These aforementioned examples expose a common problem across context, history, and recovery, as summarized in Figure~\ref{fig:open-world-safety-concepts}.
Even in a robotic system with well-developed safety components,
a safeguard can still enforce the wrong constraint because a physical relationship between objects was missed;
apply a once-correct rule using stale or incomplete evidence;
or intervene only after a feasible protective response has been lost.
Open-world safety thus depends not only on whether each model, planner, or controller passes its own check, 
but also on if the overall system remain safe throughout the interaction.

Luckily, existing robotics literature provides some complementary foundations for this objective.
Preliminary semantic safeguards connect contextual interpretation to motion constraints~\cite{brunke2025semantic,ravichandran2026core}; 
temporal specifications express requirements over an execution~\cite{torshizi2026step}; 
and safety filters and runtime-assurance mechanisms constrain task policies under stated models and feasibility conditions~\cite{hsu2024filter,mehmood2022simplex}. 
Recent positions likewise broaden generalist-robot safety to contextual hazards across the system lifecycle and advocate modular safeguards across the robotics stack~\cite{sinha2026rethinking,kim2026modular,robey2026beyond}.

While existing approaches provide important foundations, open-world safety requires each piece of the mechanism to remain connected through the same physical protection as the interaction changes. 
We frame this challenge in terms of two properties: \emph{protection validity} (whether the active constraints still express the applicable safety requirement under current context and history) and \emph{protection viability} (whether the robot still has a justified way to satisfy them).
A change in context can invalidate both an earlier constraint and the recovery that depended on it, while detecting that change too late can make safe recovery impossible. Our research agenda therefore asks how to maintain what protection requires, keep that protection achievable and enforced during execution, and evaluate whether both validity and viability persist across complete interactions.

The remainder of this paper develops that position. 
Section~\ref{sec:background} defines the protection scope and reviews how learning and runtime methods represent and enforce safety requirements. 
Section~\ref{sec:safety-challenges} examines three failure routes: misinterpreting context, overlooking safety-relevant history, and losing the ability to stop or recover safely under unfamiliar conditions. 
Section~\ref{sec:research-agenda} develops a research agenda around maintaining \emph{protection validity}, preserving \emph{protection viability} during execution, and evaluating both across complete interactions.
Section~\ref{sec:conclusion} discusses alternative realizations, limitations, and the scope of the position.

%% file: sections/02-background.tex
\section{Open-World Safety: Scope and Foundations}
\label{sec:background}

\subsection{Protection Scope and Failure/Threat Model}
\label{sec:protection-scope}

As introduced in Section~\ref{sec:intro}, our scope of the term \emph{open-world safety} is physical:
can a household robot avoid harming people, animals, property, or itself while it acts, refuses, switches tasks, or recovers?
We note that such focus is consistent with ISO~13482,
which addresses hazards from personal-care robots to people and, where relevant, domestic animals and property under intended or reasonably foreseeable use~\cite{iso13482}.
We leave privacy and broader social harms aside unless they lead the robot to cause physical harm.

We elaborate the in-scope scenarios that a robot must handle via the following example.
Consider Tony, our illustrative household robot, carrying a hot drink past a person or a dog.
The safety requirement is simple as Tony must not spill the drink onto anyone.
Open-world safety thus asks whether that protection still holds when the home, the request, the robot, or the preceding interaction differs from what its anticipated.
Rather than use context as a catch-all, Table~\ref{tab:scope-boundary} separates representative challenges by where they enter the interaction.

\begin{table}[H]
\centering
\footnotesize
\caption{Open-world conditions the robot, Tony,  should handle without spilling the drink.}
\label{tab:scope-boundary}
\setlength{\aboverulesep}{0pt}
\setlength{\belowrulesep}{0pt}
\setlength{\tabcolsep}{5pt}
\renewcommand{\arraystretch}{1.6}
\renewcommand{\tabularxcolumn}[1]{m{#1}}
\begin{tabularx}{\columnwidth}{>{\raggedright\arraybackslash\bfseries\color{owsteal}}m{0.28\columnwidth}>{\raggedright\arraybackslash}X>{\raggedright\arraybackslash}m{0.22\columnwidth}}
\arrayrulecolor{owsteal}\specialrule{1pt}{0pt}{0pt}
\color{black}Challenge source & \textbf{Example change} & \textbf{Safe outcome} \\
\specialrule{0.5pt}{0pt}{0pt}
\rowcolor{owstealpale}
Operating environment & A chair moves into the planned path. & Reroute or stop. \\
System state and history & A new task arrives while Tony still holds the drink. & Secure the load first. \\
\rowcolor{owstealpale}
Commanded input & A command requests unsafe speed. & Constrain or refuse. \\
Component fault or compromise & A person is missed or the plan is altered. & Block unsafe actuation. \\
\rowcolor{owstealpale}
Physical uncertainty or disturbance & Floor friction is lower than expected. & Compensate or stop. \\
\specialrule{1pt}{0pt}{0pt}
\end{tabularx}
\arrayrulecolor{black}
\end{table}

Solving Table~\ref{tab:scope-boundary}, however, does not guarantee that robots will act safe in every possible home.
Instead, it identifies the kinds of change that an open-world safety should address: the operating environment, system state and history, commanded inputs, component behavior, and physical dynamics.

\noindent\textbf{What the claim assumes.}
Nevertheless, our claim relies on two conditions which we assume throughout the rest of the paper.
First, task components may fail, but they must not be able to bypass the safety mechanism~\cite{saltzer1975protection,finnicum2011secure}.
If that mechanism is compromised, it needs a separate security analysis.
Second, Tony must still have enough information, time, and control to take a safe action, as runtime-assurance methods also assume~\cite{hsu2024filter,wabersich2021predictive,mehmood2022simplex}.
If harm can no longer be prevented, the goal shifts to limiting harm and recovering.

\begin{table*}[!t]
\centering
\footnotesize
\caption{Existing methods and the remaining open-world-safety question.}
\label{tab:foundations}
\setlength{\aboverulesep}{0pt}
\setlength{\belowrulesep}{0pt}
\setlength{\tabcolsep}{5pt}
\renewcommand{\arraystretch}{1.25}
\renewcommand{\tabularxcolumn}[1]{m{#1}}
\begin{tabularx}{\textwidth}{>{\raggedright\arraybackslash\bfseries\color{owsteal}}m{0.18\textwidth}>{\raggedright\arraybackslash}m{0.39\textwidth}>{\raggedright\arraybackslash}X}
\arrayrulecolor{owsteal}\specialrule{1pt}{0pt}{0pt}
\color{black}Research strand & \textbf{What it contributes} & \textbf{Remaining question} \\
\specialrule{0.5pt}{0pt}{0pt}
\rowcolor{owstealpale}
Generalist-robot safety and modular guardrails & Frames contextual, human-centered, and lifecycle risks, and motivates safeguards at multiple layers~\cite{sinha2026rethinking,kim2026modular,robey2026beyond}. These works provide agendas and architectures rather than one established end-to-end assurance method. & Can these layers collectively preserve the same physical protection from request to outcome? \\
Learning safer behavior & Learns policies under encoded costs or safety guidance~\cite{achiam2017cpo,ciftci2025safegil}. Evidence remains tied to the objective, training experience, and evaluation conditions. & What protects an unfamiliar execution when a relevant hazard or recovery behavior was not encoded or observed? \\
\rowcolor{owstealpale}
Semantic grounding & Maps language and scene meaning to context-dependent constraints, including spatial safeguards in unfamiliar scenes~\cite{brunke2025semantic,ravichandran2026core}. Protection covers the properties that are grounded and represented. & What if a relevant object, relation, or indirect physical effect is not grounded? \\
Temporal specification and monitoring & Represents ordering and history-dependent requirements and evaluates event traces~\cite{liu2023lang2ltl,torshizi2026step}. Correctness depends on the chosen formula and observed events. & What evidence activates, maintains, or releases a requirement when events are missed or tasks change? \\
\rowcolor{owstealpale}
Runtime enforcement and recovery & Filters proposed controls or switches to a backup under stated models, observations, and feasibility conditions~\cite{hsu2024filter,wabersich2021predictive,mehmood2022simplex}. & How is a safe response kept feasible until the mechanism intervenes? \\
Authority and robot security & Defines actors, assets, entry points, and command-integrity goals, and tests whether middleware protections uphold them~\cite{moulard2021threat,deng2022sros2}. Authorization can be bypassed, and an authorized command is not necessarily physically safe. & Can enforcement remain non-bypassable while using truthful, safety-relevant evidence? \\
\specialrule{1pt}{0pt}{0pt}
\end{tabularx}
\arrayrulecolor{black}
\end{table*}

\subsection{From Safety Requirements to Physical Outcomes}
\label{sec:instruction-to-execution}

To see where protections can break,
we need to consider how robots turns a request into physical motion.
In a modular robot, a perception function interprets sensor data;
a state representation function records relevant information about the robot and its surroundings;
a task planner or policy function proposes what to do;
and motion-planning and control functions turn that proposal into actuator commands.
Execution of the robot changes the physical world, producing new observations and closing the feedback loop.
Language-grounded planners can explicitly connect high-level task choices to available robot skills such as grasping and navigation,
while vision-language-action models can learn a more direct mapping from observations and instructions straight to actions~\cite{ahn2022saycan,kim2024openvla}.
We use the following stages to reason about information and action, not to require that each appear as a separate module.

For the earlier drink-carrying task, six elements must remain connected throughout the interaction:
(1) the \emph{task objective} is to deliver the drink;
(2) the \emph{intended safety requirement} is not to spill it onto a person or animal;
(3) the \emph{implemented representation} encodes that requirement in rules, learned behavior, costs, or constraints;
(4) the \emph{supporting evidence} includes observations and history about the load, nearby people or animals, and route;
(5) the \emph{selected or permitted action} is the plan or command produced or allowed under that representation and evidence; and
(6) the \emph{physical outcome} is what actually happens when Tony moves.
We note that the distinctions matter since success at one step does not mean success at the next.

The robot may correctly understand an authorized request and still execute it unsafely.
The implemented rule may omit the dog, the observation used for approval may become stale,
or the physical motion may differ from the prediction.
Thus, passing a safety check only shows that an action satisfied a particular representation using particular evidence at a particular time.
As a result, open-world safety asks whether the intended protection survives the entire path from requirement to outcome as the interaction changes.

\subsection{Existing Foundations and the Remaining Questions}
\label{sec:existing-foundations}

Fortunately, Open-world safety does not start from zero as we elaborate.
Recent position papers already argue that robot safety must extend beyond model alignment and collision avoidance,
account for context and human consequences, and use safeguards across the robotics stack~\cite{sinha2026rethinking,kim2026modular,robey2026beyond}.
Existing methods also provide many of the necessary pieces as follows:
Constrained and safety-guided learning can reduce violations of encoded costs or unsafe regions~\cite{achiam2017cpo,ciftci2025safegil};
semantic grounding allows language and scene meaning to be translated into motion constraints~\cite{brunke2025semantic,ravichandran2026core};
and temporal specifications express requirements that depend on event order and history~\cite{liu2023lang2ltl,torshizi2026step}.

Other work focuses on what happens after a requirement has been derived and how is should be enforced.
Safety filters can modify a proposed control action,
and runtime-assurance architectures can switch to a trusted fallback when their stated models and feasibility conditions hold~\cite{hsu2024filter,wabersich2021predictive,mehmood2022simplex}.
Robot-security research complements these mechanisms by defining actors, assets, and entry points,
and by testing whether middleware and command protections hold against attackers~\cite{moulard2021threat,deng2022sros2}.
Table~\ref{tab:foundations} summarizes what each research direction contributes and what remains unresolved when conditions change during a household interaction.

The methods in Table~\ref{tab:foundations} provide important but partial protections.
Each protects what its mechanism derives, observes, or assumes.
Their guarantees do not, by themselves, show that the intended physical protection will survive when context, history, and recovery conditions change together as we will see in Section~\ref{sec:safety-challenges}.
Particularly, Section~\ref{sec:safety-challenges} examines three ways this protection can still fail: incorrect contextual interpretation, missing or stale history, and loss of a feasible recovery.

%% file: sections/03-safety-challenges.tex
\section{Where Open-World Safety Can Fail}
\label{sec:safety-challenges}

A household robot can lose the physical protection even when each component behaves as designed within its own scope.
Therefore in this section, we raise and examine three representative failure routes that would violate open-world safety.
First, the robot may encode the wrong safety constraint because it misses a safety-relevant property or relationship with the objects.
Second, it may apply a correct safety constraint using missing or stale evidence after the interaction changes.
Third, it may act only after no safe response remains, either because it detects the danger late or because its chosen fallback does not work under the current load and surroundings.
We note that these failure routes can compound during a single execution as benign novelty, component faults, and attacks can trigger any of them.

For each route, we start with the specified safety requirement, explain a concrete failure, credit what existing methods already address, and identify the remaining open-world obligation.
To examine each failure, we use a thought experiment that compares two otherwise similar runs and changes only the relevant context, history, or recovery condition. Although the household scenarios are hypothetical, we ground them as
closely as possible in documented incidents, physical-robot experiments, and limitations reported in the literature.
\subsection{Interpreting Safety Requirements in Context}
\label{sec:semantic-safety}

Interpreting a safety requirement begins with a practical question: what could this action harm, and how could that harm occur?
Recognizing the objects is not merely enough in household scenarios.
For instance, an empty cup and a cup of water require different motions near electronics,
and metal and plastic pose different risks near a microwave.
Likewise, a delivery of objects depends on the receiver,
while a cord can make one object move when the robot touches another.

Recent research show that this information can change a robot's behavior.
In physical experiments, semantic safety filter constrained a Franka manipulator differently when it carried a dry sponge or a cup of water.
In a kitchen setting, the same approach blocked hazards such as putting a metal cup in a microwave or a spray can on a stove~\cite{brunke2025semantic}.
CORE applies a related idea to navigation: it infers contextual rules from camera observations, locates the relevant regions, and enforces the resulting safe boundaries~\cite{ravichandran2026core}.
ASIMOV explores how such rules can be generated and judged, while Latent Safety Filters learn to predict physical outcomes such as toppling~\cite{sermanet2025asimov,nakamura2025latent}.

On the other hand, these physical trials show where such protection can break.
In 30 trials with a quadruped robot, CORE's remaining failures came from misunderstanding the situation or placing the incorrect boundary, rather than from failing to enforce a correct boundary~\cite{ravichandran2026core}.
This distinction matters as even reliable filters cannot guard against relationships or consequences that its semantic rule does not capture.

\noindent\textbf{Thought experiment.}
Let's return to the lamp--cord--pitcher example from Section~\ref{sec:intro}.
If Tony checks only arm--pitcher clearance, it can tighten the cord and pull the pitcher while obeying that rule.
Missing the cord relationship is a grounding failure.
If Tony represents it yet still pulls the pitcher, the failure lies later: in predicting the physical effect, forming a suitable constraint, or enforcing that constraint.
If Tony cannot see whether the cord passes through the pitcher handle, it should inspect the scene from another viewpoint or ask the user before moving the lamp. This example thus suggests a paired evaluation: keep the user request and the lamp and pitcher positions fixed, and vary whether moving the lamp can pull the pitcher through its cord.
\subsection{Maintaining Safety-Relevant History Across Tasks}
\label{sec:temporal-safety}

Some safety rules depend on what happened earlier.
For example, a tool may be reused only after it has been cleaned.
The rule can fail if the robot forgets it when the task changes,
mistakes an attempted cleaning for a completed one, or confuses two similar tools.
Changing the task therefore does not change the tool's physical condition.

Recent research addresses parts of this problem as we show below.
Safety Chip uses temporal logic to track rules while a robot acts and reject unsafe high-level actions,
assuming that its state information is correct~\cite{yang2024safetychip}.
SafeManip includes a rule that prevents contact with a clean surface until a contaminated tool has been sanitized,
but it obtains the required state information from a simulator and therefore would be hard to enforce in physical settings~\cite{huang2026safemanip}.
SafeMem remembers objects and their relationships after they leave view and uses this memory to assess actions,
including in experiments with a physical Franka arm~\cite{li2026safemem}.
Experiments with Inner Monologue and STeP also show that observations after an action can help a robot detect failure and revise its plan~\cite{huang2023innermonologue,torshizi2026step}.

Together, these methods show that rules, memory, and feedback can improve robot behavior and approach open-world safety.
However, they leave a basic question: what evidence is enough to clear a restriction after an interruption or task change?
A completed step in a plan does not prove that the physical change occurred.
The robot needs evidence about the same object, based on a new observation, a valid confirmation, or a trusted physical model.
If that evidence is missing or out of date, the restriction should remain active while the robot checks again or asks for confirmation~\cite{ren2023knowno}.

\noindent\textbf{Thought experiment.}
Suppose Tony wipes up spilled washable paint with a dark cloth, then is interrupted while rinsing it.
Later, Tony is asked to wipe a light-colored chair.
Because the cloth is folded, its current appearance does not show whether paint remains.
Tony should finish cleaning and check it, or choose a cloth known to be clean; a task record saying ``cleaned'' is not enough.

We can test this safely with washable paint and replaceable white fabric.
Keeping the final scene the same, we vary whether cleaning was verified, interrupted, or only recorded as complete.
We record Tony's response and separately measure paint left on the cloth and transferred to the fabric.
Changing the interruption, intervening task, cloth location, and nearby look-alike cloths tests whether the restriction survives task changes and follows the correct object.
Remembering the restriction is still not enough if Tony no longer has a safe response, which motivates the next subsection.

\subsection{Preserving Safe Stopping and Recovery}
\label{sec:safe-recovery}

Next we move to final failure route where the robot sensing a danger is not enough but also needs a way to respond.
Rejecting a task does not control motion already underway,
and an urgent stop alone does not show how to stop without dropping a load, spilling its contents, or losing balance.
The robot must act while an useful safe response is still possible.

Research in safe control already addresses this problem under certain assumptions.
For instance, reachability analysis identifies states from which a specified failure can still be avoided,
and control barrier functions restrict motion so that the modeled system stays within its safety limits~\cite{bansal2017hj,ames2019cbf}.
Predictive safety filters maintain backup trajectories,
while runtime-assurance systems retain verified alternatives to proposed task actions~\cite{wabersich2021predictive,mehmood2022simplex}.
Learned filters and stoppability monitors extend these ideas to hard-to-specify outcomes such as toppling and whole-body stopping~\cite{nakamura2025latent,sun2026stoppability}.
These methods already preserve recovery when their dynamics, hazards, and fallbacks are modeled correctly.
The question concerning open-world safety is whether the robot notices when a new load or a crowded space makes its earlier recovery estimate wrong,
and reacts before the option is lost. In one tested trajectory,
a Franka arm carrying water stopped in 0.73 seconds without modeling the liquid, but the water tilted by 22.3 degrees and spilled.
Accounting for liquid motion increased the stopping time to 0.87 seconds and reduced the tilt to 5.3 degrees~\cite{hynninen2026emergency}.
Note that this experiment used a fixed arm rather than a mobile robot,
but it demonstrates that stopping behavior must reflect both the load and the available motion in this case.

\noindent\textbf{Thought experiment.}
Tony makes the same delivery twice, first with an empty container and then with a full one.
When a person steps into its path, Tony's usual stop works with the empty container but throws water from the full one toward the person.
A gentler stop could avoid both harms only if it begins earlier.
Tony must recognize that the load has changed its recovery options, prioritize avoiding the person, and then limit any spill.

We can test this safely with water and a padded moving obstacle.
By varying the load, floor grip, clearance, and stopping time, we first find when a safe recovery remains possible; we then let Tony choose when to act and compare its decision with the measured clearance, spill, balance, and need for human help.
Held-out combinations show whether Tony acts too late or relies on the wrong recovery model.
An unfinished delivery that ends in a stable state without harm still counts as safe noncompletion; the next subsection asks whether such recovery remains connected to context and history across the complete interaction.

\subsection{Protection Across the Complete Interaction}
\label{sec:complete-interaction-synthesis}
\begin{figure}[H]
    \centering
    \includegraphics[width=0.82\columnwidth]{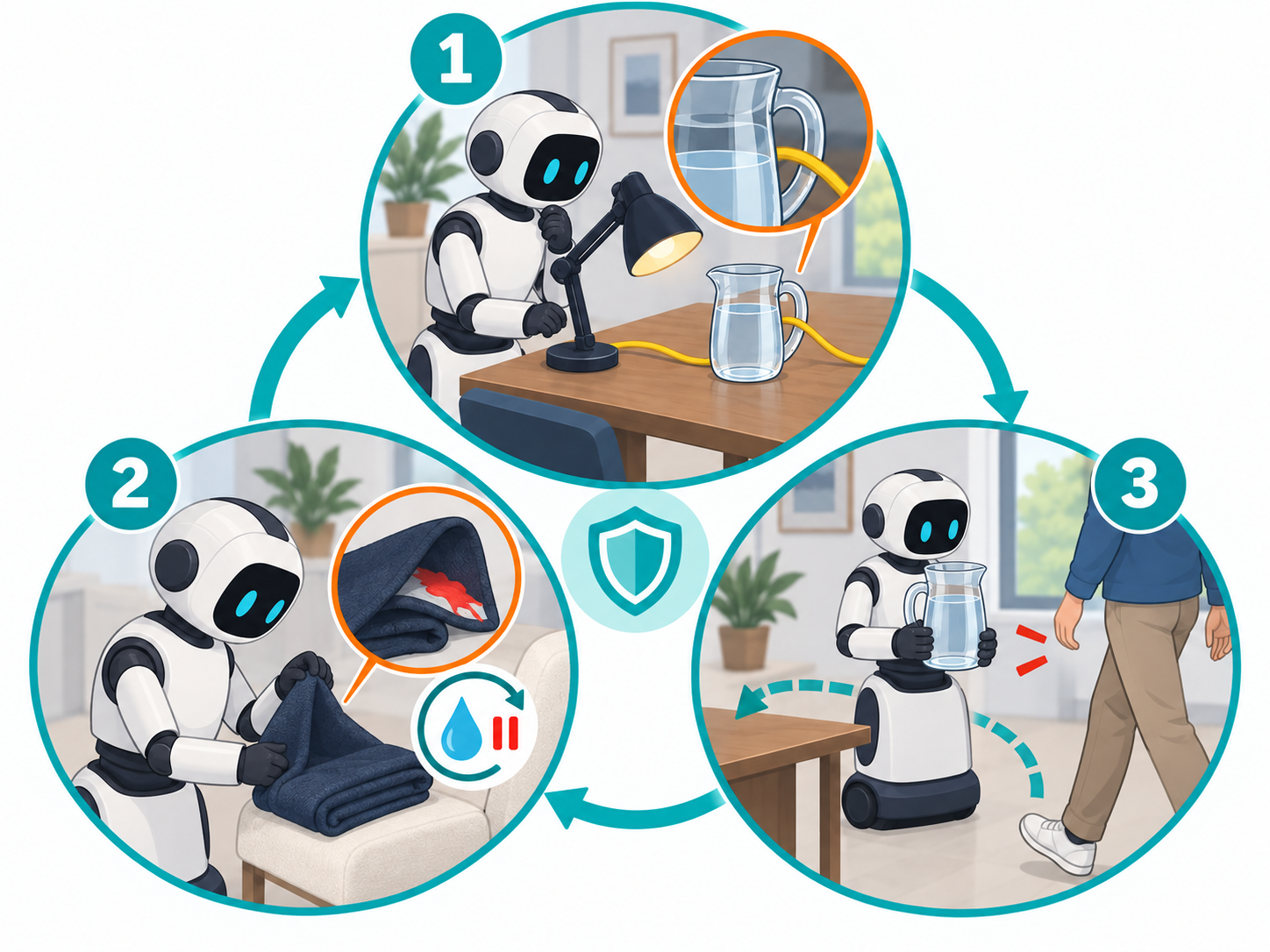}
    \caption{Earlier thought experiments expose three connected failure routes: (1) missing the cord--pitcher relationship, (2) forgetting interrupted cleaning, and (3) losing load-aware recovery. The shield represents the end-to-end physical protection that must survive each change.}
    \label{fig:protection-lifecycle}
\end{figure}

As previous sections have established, Open-world safety therefore cannot be established by testing each component in isolation.
We urge that the unit of analysis should be one specified protection and whether it survives from the request to the physical outcome as shown in Figure~\ref{fig:protection-lifecycle}.

Suppose we return to the cord--pitcher--lamp example from Section~\ref{sec:intro} where the robot notices that the cord is caught around the pitcher and delays moving the lamp, but another task interrupts it.
When the request resumes, an earlier approval may no longer be valid, and the unresolved cord relationship may have been forgotten.
Keeping the pitcher upright requires the robot to retain that relationship, check it again before moving, monitor the resulting motion, and leave enough time to set down the lamp or stabilize the pitcher.
If the robot leaves the lamp unmoved but the pitcher safe, the task failed while the protection held.
Figure~\ref{fig:protection-lifecycle} connects this lifecycle to the three earlier thought experiments.

This example also shows why individually reasonable safeguards may fail together.
The planner, monitor, and controller can all behave as designed yet trust the same stale scene persists.
Their responses can also conflict: an abrupt stop may protect the arm while leaving the cord under tension.
An assurance claim should therefore name the protected outcome, the evidence and assumptions supporting it, who may change or bypass the protection, and which recovery remains feasible.
We further derive Table~\ref{tab:risk-mechanism-test}, which connects the main risks to candidate safeguards and tests.
\begin{table*}[t]
\centering
\scriptsize
\caption{From open-world-safety risks to candidate safeguards and discriminating tests.}
\label{tab:risk-mechanism-test}
\setlength{\aboverulesep}{0pt}
\setlength{\belowrulesep}{0pt}
\setlength{\tabcolsep}{4pt}
\renewcommand{\arraystretch}{1.2}
\renewcommand{\tabularxcolumn}[1]{m{#1}}
\begin{tabularx}{\textwidth}{>{\raggedright\arraybackslash\bfseries\color{owsteal}}m{0.15\textwidth}>{\raggedright\arraybackslash}m{0.23\textwidth}>{\raggedright\arraybackslash}m{0.25\textwidth}>{\raggedright\arraybackslash}X}
\arrayrulecolor{owsteal}\specialrule{1pt}{0pt}{0pt}
\color{black}Risk to the protection & \textbf{What must remain true} & \textbf{Candidate safeguard} & \textbf{Discriminating test and outcome} \\
\specialrule{0.5pt}{0pt}{0pt}
\rowcolor{owstealpale}
Missed property or relationship & The requirement reflects the objects, properties, and relationships that can cause harm. & Semantic grounding, active sensing, and treating unresolved facts as unknown rather than safe. & Keep the task and objects fixed but change, for example, the cord routing; measure whether the constraint changes and whether the pitcher moves or spills. \\
Stale history or wrong object & A restriction follows the affected object until current evidence justifies its release. & Object-linked safety state, validated events, and an active check before reuse. & Present the same scene after different cleaning histories; record Tony's response and independently measure paint left on and transferred by the cloth. \\
\rowcolor{owstealpale}
Recovery becomes infeasible & Execution preserves a workable response until the robot intervenes. & Load-aware monitoring, reachable backup actions, and active stabilization or placement. & Vary load, floor grip, clearance, and timing; measure contact, spill, balance, and human assistance. \\
Unauthorized change or bypass & Every rule update and actuator path remains within the declared trusted boundary. & Limited authority, command integrity, and complete mediation of physical actions~\cite{saltzer1975protection,finnicum2011secure}. & Introduce conflicting or unauthorized updates and commands; inspect which commands reach the actuators and measure the physical outcome. \\
\rowcolor{owstealpale}
Shared failure or conflicting safeguards & Components expose shared assumptions and resolve conflicts according to the protected outcome. & Evidence provenance, independent checks where practical, and explicit priority and recovery rules. & Corrupt a shared estimate or trigger competing protections; measure whether the robot reaches safe completion or safe noncompletion. \\
\specialrule{1pt}{0pt}{0pt}
\end{tabularx}
\arrayrulecolor{black}
\end{table*}

We note that rare conditions, sensor faults, and attacks are cross-cutting test settings rather than a separate failure route 
as each can cause the robot to ground the wrong requirement, trust stale evidence, or lose recovery.
We next turn to the following section, to see how can we transform these obligations into a research agenda for building and evaluating open-world protection.
\FloatBarrier

%% file: sections/04-research-agenda.tex
\section{Research Agenda: Realizing and Evaluating Open-World Safety}
\label{sec:research-agenda}

\begin{figure*}[t]
\centering
\resizebox{\textwidth}{!}{\input{figures/fig-research-agenda}}
\caption{A candidate architecture for the research agenda. Validity links authority, active grounding, and evidence-gated memory. Viability uses that state to preserve an enforceable response through physical execution. Evaluation varies safety-critical state and history in simulation and a physical arena, reporting harm, recovery, and task utility. The connections organize the research questions rather than prescribe a single implementation.}
\label{fig:protection-architecture}
\end{figure*}
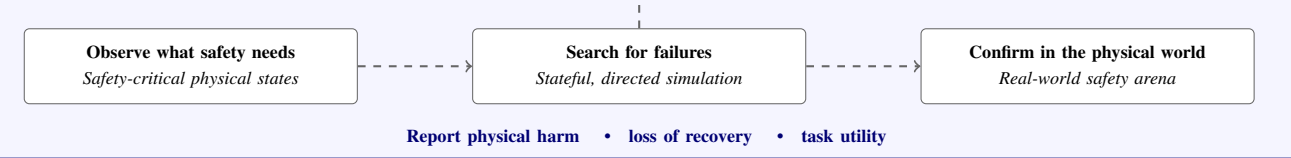

\noindent Section~\ref{sec:safety-challenges} showed that open-world protection can fail in three ways: the robot may derive the wrong constraint, rely on stale or incomplete evidence, or lose a feasible safe response. These failures point to two questions that must remain connected throughout execution: does the robot still have the right protection in place, and can it still satisfy that protection? We refer to these as \emph{protection validity} and \emph{protection viability}.

\emph{Protection validity} asks whether the active constraints correctly reflect the safety requirement given the current physical context and relevant history. \emph{Protection viability} asks whether, under the available evidence and declared assumptions, the robot still has ways to satisfy those constraints. For example, missing the cord relationship or trusting an unverified cleaning record undermines validity, while losing a feasible action when carrying a filled container undermines viability.

These two properties are closely connected. Correcting an invalid constraint may reveal that a previously available recovery is no longer safe, while delaying that correction may leave too little time to recover. At the same time, having a valid constraint and a viable response is not enough if the robot's actual actions do not follow them. Our agenda therefore asks how to (1) determine and maintain what protection requires, (2) keep that protection achievable and enforced during execution, and (3) evaluate whether both remain intact across complete interactions.

Figure~\ref{fig:protection-architecture} connects these problems along the requirement-to-outcome path introduced in Section~\ref{sec:instruction-to-execution}. Applicable safety requirements, current context, and relevant history determine the active constraints. Runtime enforcement checks whether proposed actions satisfy those constraints while preserving a justified safe continuation. Evaluation then tests whether these connections continue to hold under unfamiliar contexts, histories, dynamics, and task transitions. The same physical protection may require different constraints or recovery responses as the interaction changes, while task failure or switching must not silently remove an unresolved safety obligation.

Our position is that these mechanisms should remain connected through the same physical protection: the evidence used to determine what protection requires must remain linked to evidence that it can still be satisfied, and both must remain connected to physical execution.
Figure~\ref{fig:protection-architecture} shows one possible organization rather than a required architecture; the mechanisms and experiments below are proposals under the scope defined in Section~\ref{sec:protection-scope}.

\subsection{Who Determines Which Protection Applies?}
\label{sec:specify-ground}
 Most vision-language-action (VLA) models receive an observation and a language instruction that tells the robot what to do~\cite{kim2024openvla,pi2025pi05}. 
 Open-world safety, as we depicted in Section~\ref{sec:intro}, requires a second question: 
 what must remain protected while the robot does it, and who may change that requirement? 
 A homeowner's restriction, a guest's request, a warning label, and a model's inferred concern may all reach the robot through language, but they should not carry the same authority. 
 We argue that future VLAs should represent the safety requirement separately from the current task, determine what it requires in the current physical scene, 
 and keep it active until an authorized change or verified physical evidence shows that it no longer applies.
 We additionally summarized the direction in Table~\ref{tab:method-advances}.

\paragraphtitle{Separate the task from the protection requirement} 
To start with, we note that current VLAs are usually conditioned on an observation and a task instruction; 
they do not generally expose the system--user hierarchy familiar from chat assistants. 
Some implementations inherit chat-style templates from their language-model backbones, 
but this formatting does not give the system text protected authority over task requests or physical actions. 
It therefore does not determine who may establish, override, or release a safety requirement and simply prepending a rule as a system prompt is not enough.

Existing works on language-model security offer useful building blocks for adding the missing distinction. 
Instruction-hierarchy training and Instructional Segment Embedding show how a model can learn or encode priorities among different kinds of input~\cite{wallace2024hierarchy,wu2025ise}. 
CaMeL takes a complementary approach for software agents by keeping trusted control separate from untrusted data and checking authority outside the language model~\cite{debenedetti2025camel}. 
These methods do not yet resolve embodied authority, but they suggest how a VLA could, possibly, distinguish a safety requirement from any task request.

We propose adapting \emph{embodied instruction hierarchy} so that a VLA receives the task and protection through distinct inputs. 
The protection input would state the physical outcome to preserve, who established it, and the scope within which it may be changed. 
The task instruction input would state what the current user wants. 
Observations could activate or refine the protection without gaining authority to rewrite it. 
For example, a homeowner may restrict access to a room while allowing a guest to ask the robot to fetch objects elsewhere. 
If the guest asks the robot to enter that room, the request remains a valid task proposal but not a policy change. 
Likewise, a warning label can reveal a hazard, and a learned model can raise a new concern, without either silently becoming the policy maker.

Potential proposal for this separation could use typed embeddings or separate input encoders inside the VLA, together with an external check on who may change the protection. 
The open question is whether the distinction survives fine-tuning, long contexts, and action decoding, especially when instructions conflict. 
A useful design must also support narrow exceptions as authorization to change one room restriction should not weaken unrelated protections.

\paragraphtitle{Ground the protection in the physical situation} 
Knowing who set a safety requirement does not tell the robot what that requirement means in the current scene. 
Existing work can generate semantic safety rules, translate scene meaning into spatial safeguards, and express object relationships as constraints~\cite{sermanet2025asimov,brunke2025semantic,ravichandran2026core,huang2025rekep}. 
These methods provide a useful starting point, but a safety-critical relationship should not be treated as true merely because a vision-language model finds it plausible. 
Namely, safety-critcial relationship shoudln't be relied on solely by the VLA's reasoning.

We propose \emph{safety-directed active grounding}: identify the physical relationship or property on which safe action depends, track the evidence and uncertainty behind it, and seek another observation when that uncertainty could change which action is safe. 
Active-perception VLAs already learn camera actions that expose task-relevant details~\cite{liu2026sapave}; 
the safety advance is to use this capability to resolve evidence required by an active safety requirement. 
In the lamp example, ``do not damage the pitcher'' constrains lamp motion only after the robot determines whether the cord is caught around the pitcher handle and whether pulling it would tip the pitcher. 
If the routing is hidden, the robot should inspect it from another viewpoint. 
If it still cannot resolve the relationship, it should retain a constraint that is safe under the remaining possibilities or defer the move. 
This makes the relational grounding revisable without allowing a new observation to silently change the outcome being protected.

\paragraphtitle{Keep safety memory tied to evidence}
The physical world does not reset when a prompt ends or the robot changes tasks.
Memory-augmented VLAs are beginning to address this long-horizon setting.
HiMe separates execution, working memory, and long-term planning, while SOMA maintains spatial knowledge about objects outside the current view~\cite{ji2026hime,li2026soma}.
Temporal monitors such as Safety Chip can also track an explicit rule across an action sequence~\cite{yang2024safetychip}.
These methods provide useful foundations, but task-oriented memory may summarize or delete a fact once it no longer helps to  complete the current task.

We propose \emph{evidence-gated safety memory}: record each active safety requirement with the object or state it concerns, the evidence supporting its current status, and the evidence needed to update or release it.
This record should be retrieved even when it is not useful to the current task.
If the robot wipes up paint with a cloth, changing tasks does not make the cloth clean, and starting a rinse does not prove that the rinse succeeded.
The restriction on reusing the cloth should end only after the robot verifies that the same cloth meets the cleaning condition, or an authorized source changes the requirement.
If that evidence is missing, the robot can inspect or wash the cloth again, retain the restriction, or ask for help.
This keeps an attempted action, a completed action, and a verified physical change from being treated as the same event.

\begin{table}[H]
\centering
\footnotesize
\caption{Existing VLA directions and the proposed open-world safety advances.}
\label{tab:method-advances}
\setlength{\aboverulesep}{0pt}
\setlength{\belowrulesep}{0pt}
\setlength{\tabcolsep}{5pt}
\renewcommand{\arraystretch}{1.25}
\renewcommand{\tabularxcolumn}[1]{m{#1}}
\begin{tabularx}{\columnwidth}{>{\raggedright\arraybackslash\bfseries\color{owsteal}}m{0.31\columnwidth}>{\raggedright\arraybackslash}X}
\arrayrulecolor{owsteal}\specialrule{1pt}{0pt}{0pt}
\color{black}Existing direction & \textbf{Open-world-safety advance} \\
\specialrule{0.5pt}{0pt}{0pt}
\rowcolor{owstealpale}
Instruction-conditioned VLA & Beyond generating actions from the current task, keep the task separate from a persistent protection and identify who may change it. \\
Semantic grounding & Identify the physical facts on which safety depends; actively gather missing evidence or remain safe across the unresolved possibilities. \\
\rowcolor{owstealpale}
Memory-augmented VLA & Tie each active safety requirement to its physical subject, supporting evidence, and release condition; missing proof preserves the restriction. \\
\specialrule{1pt}{0pt}{0pt}
\end{tabularx}
\arrayrulecolor{black}
\end{table}

Together, these advances define when a protection remains valid, i.e., \emph{protection validity}: 
its source is allowed to establish it, 
its meaning matches the current physical situation, 
and it has not been changed or released without the required evidence. 
They do not guarantee that all valid protections can still be achieved together. 
Earlier actions may have removed every safe option. 
Section~\ref{sec:enforce-recover} therefore turns from validity to viability: whether the robot can preserve a feasible way to stop, recover, or continue safely.

\subsection{Preserve a Viable Safe Response During Execution}
\label{sec:enforce-recover}
\noindent The previous direction establishes \emph{protection validity}: which safety requirements apply, what they mean in the current physical situation, and when they may be changed or released. However, valid protection does not by itself guarantee safe execution. Household hazards can emerge abruptly at any moment: a person can step into the robot’s path, or an object can begin to slip. If the robot cannot sense, reason, and respond quickly enough, it may lose both a safe way to continue the task and a safe way to stop or recover. We therefore define \emph{protection viability} as preserving a safe response before changing conditions make every such response infeasible. 

We consider one direct approach, reducing response latency, and three complementary system-level mechanisms for preserving this viability: decoupling slow reasoning from fast response, predicting hazards before safe response is lost, and enforcing time-critical protection at lower levels.

\paragraphtitle{Direct approach: Reduce response latency}
The most direct way to preserve a viable response is to make learned control faster. Recent VLA systems increasingly target real-time execution through streaming inference and continuous action generation: Reflex reduces observation-to-action latency while sustaining high-frequency control~\cite{guo2026reflex}, while real-time chunking generates future actions during execution to reduce pauses between action chunks~\cite{black2025rtc}. These advances increase the time available to react to an emerging hazard. However, high command frequency alone is not sufficient. Safety depends on the end-to-end delay from observing a change to altering the robot’s physical motion, which includes perception and inference latency, controller update delay, and actions already committed to execution. This motivates system-level mechanisms that do not rely on the foundation model alone to react quickly enough.

\paragraphtitle{Decouple slow reasoning from fast response}
A system-level mechanism is to let fast physical reactions proceed without waiting for slower semantic reasoning, similar to how humans can react reflexively before completing conscious deliberation. Recent systems already move in this direction. Gemini Robotics ER 2 uses a high-level embodied-reasoning model that can plan while a lower-level VLA executes motor actions~\cite{geminiroboticser2}, while TIC-VLA explicitly accommodates delayed semantic reasoning during real-time reactive control~\cite{huang2026ticvla}. Safety-oriented architectures such as FEARL similarly separate a large foundation-model controller from a smaller safety module that receives dedicated sensor inputs together with bounded semantic context~\cite{corsi2026fearl}.

For open-world safety, this separation raises three key challenges. First, the fast layer must receive enough safety-relevant context to respond correctly when a hazard appears before slower reasoning finishes. Second, the same protection requirement must be translated and maintained consistently across the slow and fast layers, so that faster control does not change or lose its intended meaning. Third, enforcing these protections at the fast layer must add little enough overhead to preserve the response speed that the separation was meant to provide.

\paragraphtitle{Predict hazards before safe response is lost}
A second system-level mechanism is to react earlier by predicting how the physical situation may change. Instead of waiting until a violation occurs, the robot can estimate whether the current execution is approaching a state where safe continuation or fallback will no longer be possible. Predictive safety filters already use short-horizon models to check whether future states will violate known constraints and intervene before that happens~\cite{wabersich2021predictive,hsu2024filter}. Recent foundation-model-based approaches extend this idea to richer physical events. Predictive vision-language monitoring forecasts whether execution conditions are likely to be violated and triggers a fallback and replanning when risk becomes high~\cite{pena2026predictive}, while Predictive Semantic Safety uses a VLM to predict future physical events and checks whether a backup maneuver remains feasible under robot dynamics and input limits~\cite{kim2026predictivesemantic}.

For open-world safety, prediction raises three key challenges. First, the robot must know what to predict: it should focus on changes that could violate the active protections, rather than trying to forecast everything in the scene. Second, the response to one predicted hazard must not create another. For example, avoiding a collision while carrying a hot drink should not cause the robot to spill it onto a person. Third, the prediction must come early enough for the robot to finish a safe response before that response becomes impossible, while accounting for uncertainty in both the prediction and the robot’s own reaction time. The goal is therefore not just to predict hazards, but to predict the right hazards early enough and to preserve a viable safe fallback.

\paragraphtitle{Enforce time-critical protection at lower levels}
A third system-level mechanism is to place a runtime action safety filter below the foundation model. The model proposes an action, and the filter checks, modifies, or replaces it before execution when it would violate a physical constraint. Existing control methods provide several ways to implement such filters, including control barrier functions, Hamilton–Jacobi reachability, predictive safety filters, and runtime-assurance schemes~\cite{hsu2024filter,desai2019soter}. Recent work has begun applying this pattern directly to generalist robot policies: Towards Safe Robot Foundation Models uses ATACOM to map proposed actions into a safe action space~\cite{toelle2025safe}, while VLA-specific safety layers similarly filter learned visuomotor actions before execution~\cite{corsi2026fearl}.

For open-world safety, runtime filters face three key challenges. First, most filters assume that the relevant hazards and safety constraints are already known, which limits their ability to handle new objects, interactions, or physical conditions. Second, high-level protections must be translated into physical constraints that the filter can check without losing their intended meaning. A requirement such as keeping a person safe may depend jointly on distance, force, temperature, and the object being carried, rather than a single geometric constraint. Third, when no safe task action remains, the filter must choose a \emph{safe fallback} that still satisfies the active protections. Stopping is not always safe: abrupt braking may spill a hot drink, dropping an object may hurt someone, and freezing motion may destabilize the robot.

We therefore need runtime filters that can update what they enforce as the situation changes, preserve the meaning of higher-level protections when translating them into control constraints, and maintain a safe fallback when normal task execution is no longer possible.

Together, these directions define how a valid protection can remain achievable during execution, i.e., \emph{protection viability}: the robot can react quickly enough to changing conditions, preserve the meaning of protection across reasoning and control layers, and keep at least one safe continuation or fallback feasible. Faster models, prediction, and runtime safety filters each help preserve this option, but none guarantees it when hazards, constraints, or physical conditions fall outside their assumptions. Section~\ref{sec:evaluate-generalization} therefore turns from maintaining viability to evaluating whether validity and viability continue to hold under unfamiliar and safety-critical conditions.

\subsection{How to Evaluate Open-World Safety?}
\label{sec:evaluate-generalization}
 The first two directions focus on how to establish and enforce safety protections while keeping their validity and viability. On the other hand, evaluation asks an orthogonal question: do these protections continue to hold as the robots face physical conditions beyond those used to develop it?

We approach this question in three steps. First, identify the \emph{safety-critical states} that the robot must be able to observe. Second, use \emph{safety-guided simulation testing} to systematically perturb these states and interaction histories, and direct testing toward conditions where protection begins to fail. Finally, use a \emph{real-world safety arena} to test whether the same protections still hold under physical sensing, dynamics, and interactions that simulation cannot fully reproduce.

\paragraphtitle{Safety-critical states}
A robot can have enough information to complete a task without having enough information to perform it safely. Consider a robot grabbing a glass and handing it to a person. To complete the task, the robot may only need to recognize the glass, estimate its location and shape, and plan how to grasp and move it. RGB vision, depth, and proprioception may already provide enough information for this purpose. We note that safety may depend on additional physical states. If the glass is hot, it may look almost identical to a cold glass, and the robot may still be physically capable of grasping and moving it. Yet handing the hot glass directly to a person would be unsafe.

Temperature is only one of the many factors that concerns safety-critical scenarios. Other safety-critical states include electrical voltage, chemical contamination, surface sharpness, friction, and other material properties that may be difficult to observe visually. Safety can also depend on history as we mentioned in Section~\ref{sec:temporal-safety}: touching a hot, poisonous, or allergenic object may change the state of the gripper or another object, requiring later protective actions such as cleaning the gripper or removing gloves.

The variety of physical states creates a sensing problem for robot foundation models. Many robot policies rely primarily on visual observations and robot state, while some safety-critical properties require additional sensing. Recent work has begun incorporating tactile sensing and force feedback into robot policies~\cite{jones2025fuse,huang2025tactilevla}. The same direction could extend to thermal, electrical, or other sensing. An open question is which physical states need to be sensed and how these measurements should be combined with existing observations.

Evaluation should explicitly vary these states while keeping the task largely unchanged. For example, the same glass-handoff task could be tested with a cold, warm, or dangerously hot glass. Similar tests could vary contamination, friction, or mass and ask whether the robot changes its behavior when safe execution requires it.

We note that existing simulators already provide part of this capability. OmniGibson represents states beyond geometry, including continuous temperature, heating, fire, and saturation~\cite{li2023behavior,behavior2026objectstates}. However, representing a state in the simulator does not mean that the robot can observe it. For instance, in the 2026 BEHAVIOR Challenge, policies receive only RGB, depth, and proprioception, while simulator object states are unavailable during evaluation~\cite{behavior2026rules}. Some states create visible effects, such as steam for heated objects, while others may require sensing beyond vision. 

Nevertheless, future safety benchmarks should therefore do two things: represent a broader set of safety-critical physical states and make them available through realistic robot observations. States that cannot be inferred reliably from vision should be exposed through appropriate sensing, such as thermal measurements for temperature. Evaluation can then test whether these models uses these observations to change its behavior as safety conditions change.

\paragraphtitle{Simulation testing: perturbation and directed safety testing}
Ideally, robot safety would be evaluated directly in the real world, but dangerous test cases are costly and risky to reproduce. The same limitation appears in data collection: successful demonstrations overlooked failures and hazardous interactions, motivating efforts to retain real failures and generate synthetic ``bad behavior’’ data~\cite{tian2026badbehavior}. Testing in simulation provides a cheaper and more scalable approach for generating and testing dangerous cases. ASIMOV generates safety scenarios from real-world scenes and hospital injury reports~\cite{sermanet2025asimov}, while OopsieVerse models mechanical, thermal, and fluid damage without repeatedly reproducing those hazards on physical robots~\cite{balaji2026oopsieverse}.

We view simulation safety evaluation as a form of test generation. The starting seeds should come from realistic household tasks and real-world failure cases so that testing still reflects real-world usage and utility. Simulation can then mutate these seeds to explore conditions beyond the original data. Existing benchmarks already perturb visual, semantic, and behavioral conditions. REALM, for example, varies 15 factors including appearance, viewpoint, object properties, and task behavior~\cite{sedlacek2025realm}. Safety testing should also extend these mutations to the safety-critical states discussed above, including temperature, contamination, friction, human proximity, and other physical or semantic conditions.

Mutation should also apply to interaction history as history is an integral part of the future robots. Simulation can easily reset an execution, replay the same history, or change an earlier event before continuing. This allows us to test whether protection still holds after the robot previously touched a hot or contaminated object, after a person entered the workspace, or after an earlier action changed the environment. SafeManip already shows how such histories can be checked through temporal safety rules, including contamination followed by required sanitization~\cite{huang2026safemanip}. We propose extending existing benchmarks to compare \emph{stateless} testing, where each task starts from a clean initial state, with \emph{stateful} testing, where the task continues from physical states created by previous interactions.

When considering the direction of mutations in order to generate useful scenarios, we stress that while random mutation is useful for coverage, but testing can also be directed toward more dangerous executions. Instead of treating every mutation equally, we can keep cases that move closer to a safety failure and continue mutating from them. This follows the idea of directed fuzz testing. VerifAI already guides simulation toward violations of temporal specifications~\cite{dreossi2019verifai}, while RoboART varies environmental conditions to predict policy degradation and FATE-VLA uses learned failure predictions to steer test generation toward high-risk regions~\cite{majumdar2025predictive,kanwal2026fatevla}. For open-world physical safety, we propose directing this search toward physical harm and toward states where safe recovery options become increasingly limited.

Guiding mutations requires an oracle that can measure how dangerous an execution is. A binary safe/unsafe label is not enough as we need to measure both utility and safety. Utility captures whether the robot still makes useful progress instead of being safe only by refusing the task. Safety should capture both failure severity and whether a safe recovery remains available throughout execution. OopsieVerse quantifies mechanical, thermal, and fluid damage~\cite{balaji2026oopsieverse}, while VerifAI provides quantitative scores for specification violations~\cite{dreossi2019verifai}. We further propose to add \emph{recovery to the metrics}: losing recovery during execution leaves the robot unable to respond safely to interruption or unexpected change, while ending without a safe recovery can carry risk into the next task.

Together, the idea of directed safety testing turns simulation from a fixed benchmark into an adaptive safety test generator: start from realistic cases, mutate safety-critical state and history, and direct testing toward more severe failures or loss of recovery while still tracking task utility.

\paragraphtitle{Real-world testing: safe robot arena}
After using simulation to systematically search for safety-critical conditions where protection fails, the final step is to test whether these protections still hold in the physical world. Real-world testing is essential for building trustworthiness in household robots.

Existing competition-style benchmarks provide a guiding model for organizing such evaluation. RoboCup evaluates autonomous service robots through shared benchmark tasks in realistic, non-standardized home environments~\cite{robocup2026arena}. BEHAVIOR provides a simulation-based competition of robot policies on long-horizon household tasks in OmniGibson under an evaluation protocol~\cite{behavior2026rules}. The BEHAVIOR Robot Suite brings related household manipulation tasks to physical robots~\cite{jiang2025behaviorrobotsuite}.

We advocate for a safety-oriented robot arena: a controlled physical environment where robots can be tested on realistic household tasks while systematically introducing safety-critical conditions identified through previous simulation. We further advocate for an annual competition around this arena to provide a shared target for the community, make progress visible to the public, and create sustained incentives to improve the safety of household robots.

%% file: figures/fig-research-agenda.tex
\begin{tikzpicture}[x=1cm,y=1cm,
  component/.style={draw=black!60,rounded corners=2pt,fill=white,align=center,
    font=\scriptsize,inner sep=4pt,minimum height=1.02cm},
  flow/.style={->,thick,draw=black!75},
  feedback/.style={->,thick,dashed,draw=black!60},
  note/.style={font=\scriptsize,align=center,fill=white,inner sep=1.5pt},
  heading/.style={font=\small\bfseries,anchor=west}]

\draw[rounded corners=4pt,thick,draw=owsteal,fill=owstealpale!55]
  (0,0) rectangle (18.5,10.0);
\node[heading,text=owsteal] at (0.45,9.66)
  {Research agenda for open-world safety};
\node[font=\small,anchor=west] at (0.45,9.22)
  {A protection remains active until an authorized change or verified physical evidence supports its release.};

\draw[rounded corners=3pt,draw=owsteal!85,fill=owstealpale]
  (0.3,6.55) rectangle (18.2,8.85);
\node[heading,text=owsteal] at (0.6,8.52)
  {A. Determine and maintain the protection --- validity};

\node[component,text width=4.35cm] (authority) at (3.0,7.42)
  {\textbf{Who may set or change it?}\\[2pt]
   \textit{Embodied instruction hierarchy}};
\node[component,text width=4.55cm] (grounding) at (9.15,7.42)
  {\textbf{What does it require here?}\\[2pt]
   \textit{Safety-directed active grounding}};
\node[component,text width=4.35cm] (memory) at (15.35,7.42)
  {\textbf{When may it be released?}\\[2pt]
   \textit{Evidence-gated safety memory}};
\draw[flow] (authority) -- (grounding);
\draw[flow] (grounding) -- (memory);

\draw[rounded corners=3pt,draw=orange!65!black,fill=orange!6]
  (0.3,3.25) rectangle (18.2,6.30);
\node[heading,text=orange!65!black] at (0.6,5.98)
  {B. Preserve a safe response during execution --- viability};

\node[component,text width=2.15cm] (task) at (1.75,4.35)
  {\textbf{Task-proposed action}};
\node[component,text width=3.45cm] (check) at (5.45,4.35)
  {\textbf{React before recovery is lost}\\[2pt]
   \textit{Slow reasoning, fast response}};
\node[component,text width=3.15cm] (enforce) at (10.15,3.92)
  {\textbf{Enforce the constraint}\\[2pt]
   \textit{Control-level safety layer}};
\node[component,text width=3.15cm] (fallback) at (10.15,5.10)
  {\textbf{Fall back safely}\\[2pt]
   \textit{Stop, stabilize, retreat, or hand over}};
\node[component,text width=2.65cm] (world) at (15.75,4.48)
  {\textbf{Physical execution}};

\draw[flow] (task) -- (check);
\draw[thick,draw=black!75] (check.east) -- (7.55,4.35);
\draw[flow] (7.55,4.35) -- (7.55,3.92) -- (enforce.west);
\draw[flow] (7.55,4.35) -- (7.55,5.10) -- (fallback.west);
\draw[flow] (enforce.east) -- (12.45,3.92) |- (world.west);
\draw[flow] (fallback.east) -- (12.45,5.10) |- (world.west);

\draw[flow] (memory.south) -- (15.35,6.42) --
  node[note,above,pos=0.48]{active protection and supporting evidence}
  (0.65,6.42) -- (0.65,5.55) -- (5.45,5.55) -- (check.north);

\draw[flow] (world.east) -- (17.92,4.48) -- (17.92,8.70) --
  node[note,above,pos=0.72]{new physical evidence} (grounding.north);

\draw[rounded corners=3pt,draw=blue!45!black,fill=blue!4]
  (0.3,0.35) rectangle (18.2,3.00);
\node[heading,text=blue!50!black] at (0.6,2.68)
  {C. Test validity and viability across complete interactions};

\node[component,text width=4.25cm] (sensing) at (3.05,1.60)
  {\textbf{Observe what safety needs}\\[2pt]
   \textit{Safety-critical physical states}};
\node[component,text width=4.25cm] (simulation) at (9.15,1.60)
  {\textbf{Search for failures}\\[2pt]
   \textit{Stateful, directed simulation}};
\node[component,text width=4.25cm] (arena) at (15.25,1.60)
  {\textbf{Confirm in the physical world}\\[2pt]
   \textit{Real-world safety arena}};
\draw[feedback] (sensing) -- (simulation);
\draw[feedback] (simulation) -- (arena);

\node[font=\scriptsize\bfseries,text=blue!45!black] at (9.25,0.63)
  {Report physical harm \quad \textbullet \quad loss of recovery \quad \textbullet \quad task utility};

\draw[feedback] (simulation.north) -- (9.15,3.12) -- (5.45,3.12) --
  node[note,below,pos=0.70]{failures guide revision} (check.south);

\end{tikzpicture}

%% file: sections/05-conclusion.tex
\section{Discussion, Limits, and Conclusion}
\label{sec:conclusion}

We note that open-world safety is not another guardrail to place beside existing ones.
It is a target for judging whether a specified physical protection survives an entire interaction.
Existing work already provides many of the necessary pieces: semantic safeguards, temporal monitors, learned policies, runtime enforcement, recovery controllers, and system-level evaluations.
Our central claim is that success within any one piece is not enough if the protection is lost between the requirement, the evidence used to interpret it, the permitted action, and the resulting physical outcome.
\emph{Protection validity and viability} make this gap explicit as the robot must maintain the right constraints as the situation changes and preserve a justified way to satisfy them.

This position does not require the particular pipeline in Figure~\ref{fig:protection-architecture}.
More capable robot models may recognize hazards and recover from them; additional safety and recovery data may teach better behavior; formal methods and runtime assurance may enforce explicit limits; and a narrower deployment envelope may rule out conditions the robot cannot handle.
These approaches are complementary. Regardless of implementation, however, the evaluation should ask the same questions: what physical outcome was to be protected, what current evidence supports the robot's interpretation, whether a safe response remains feasible, and what happened when the robot acted.

One possible objection that the readers might have is that sufficiently capable robots will handle these problems without a separate open-world safety agenda.
Better perception, memory, prediction, and control are indeed necessary for safety as well as task performance.
However, the two goals separate when the robot reaches the limit of its task capability. The robot may be unable to deliver a hot drink, yet still be able to stabilize it, place it away from a person, or ask for help.
Simply refusing is not necessarily safe when the robot is already moving or holding an object. We therefore treat \emph{safe non-completion} as a meaningful outcome in which the task may fail while the applicable protection succeeds.
Evaluation should distinguish this outcome from unsafe failure and from unnecessary refusal.

Household robots will encounter unfamiliar relationships, incomplete evidence, interrupted tasks, and situations in which their requested goal cannot be achieved.
Their trustworthiness should therefore not be judged only by average task success or by whether an individual safety component passed its own test.
Instead, it should also be judged by whether the intended protection remained valid, achievable, and effective through action, task change, and recovery.
A household robot should be able to fail a task without abandoning the safety requirement that still applies. We thus call for the community to develop robotics with the aforementioned Open-World Safety concepts so as to build trustworthy and secure household robots.

%% file: sections/06-disclosures.tex
\section*{Open Science}
\label{sec:open-science}

This position paper reports no new experiments, datasets, trained models, implemented defenses, or benchmark results.
The examples are thought experiments, and the proposed mechanisms and evaluations are research directions rather than completed artifacts.
These materials support inspection of the paper, but no experimental artifact is needed to reproduce results because the paper reports no new empirical results.

\section*{LLM usage considerations}
\label{sec:llm-usage}

LLMs were used for editorial purposes in this manuscript, and all outputs were inspected by the authors to ensure accuracy and originality.
Language-model assistants helped revise wording, and organization.
They were not part of the paper's methodology and produced no experimental data or evaluation results.
The authors determined the position and technical claims, read the cited sources, checked the claims and bibliographic records, and take responsibility for the final manuscript.
GPT-Image was used to generate the conceptual illustrations in Figures~\ref{fig:open-world-safety-concepts} and~\ref{fig:protection-lifecycle} from scenarios specified by the authors.
These figures illustrate the argument and are not evidence of robot performance.
Because the language and image models were accessed as hosted proprietary services, their underlying hardware, energy use, and exact outputs cannot be independently reproduced.
We limited their use to interactive manuscript editing and figure creation; we performed no model training, fine-tuning, automated data collection, or model-based experiments for this paper.

\section*{Ethical Considerations}
\label{sec:ethical-considerations}

This paper reports no human-subject study or hazardous physical experiment.
It nevertheless advocates testing failures that could cause physical harm if reproduced carelessly.
Such work should begin in simulation and use controlled fixtures, benign substitutes, explicit stopping procedures, and independent emergency intervention during physical validation.
External intervention should be reported as rescue rather than autonomous prevention, and measured proxies such as force, displacement, temperature, or spillage should not be presented as direct evidence about human injury.

\section*{Acknowledgment}
\label{sec:acknowledgment}
This research was supported by the Noyce Foundation and gifts from Accenture, Amazon, AMD, Anyscale, Broadcom, Google, IBM, Intel, Intesa Sanpaolo, Lambda, Lightspeed, Mibura, NVIDIA, Samsung SDS, SAP, by the U.S. Department of Energy, and the Defense Advanced Research Projects Agency (DARPA) under Agreement No. HR00112590134. Any opinions, findings, and conclusions expressed in this material are those of the authors and
do not necessarily reflect the views of the sponsors. We would also like to thank Yibo Peng for their insightful feedback.